\documentclass{article}

\usepackage{microtype}
\usepackage{graphicx}
\usepackage{subcaption}
\usepackage{booktabs}
\usepackage{array}
\usepackage{tabularx}

\usepackage{hyperref}

\usepackage[preprint]{icml2026}

\usepackage{amsmath}
\usepackage{amssymb}
\usepackage{mathtools}
\usepackage{amsthm}
\usepackage{bm}

\usepackage[capitalize,noabbrev]{cleveref}

\DeclareMathOperator{\GELU}{GELU}
\DeclareMathOperator{\LN}{LN}
\DeclareMathOperator{\Drop}{Drop}
\DeclareMathOperator{\MHA}{MHA}
\DeclareMathOperator{\softmax}{softmax}
\DeclareMathOperator{\NB}{NB}
\DeclareMathOperator{\Var}{Var}
\DeclareMathOperator{\Corr}{Corr}
\DeclareMathOperator{\rank}{rank}
\DeclareMathOperator*{\median}{median}
\DeclareMathOperator{\diag}{diag}
\DeclareMathOperator{\MAE}{MAE}
\DeclareMathOperator{\AP}{AP}
\DeclareMathOperator{\softplus}{sp}
\newcommand{\vis}{\mathrm{visible}}

\makeatletter
\newcommand{\aivcaffilblock}{%
  \ificmlshowauthors
    \vskip 0.04in
    {\centering
      \small\setlength{\parskip}{1pt}
      \forloop{@affilnum}{1}{\NOT \value{@affilnum} > \value{@affiliationcounter}}{%
        \textsuperscript{\arabic{@affilnum}}\csname @affilname\the@affilnum\endcsname\par}%
      \vskip 0.06in
      \textsuperscript{*}These authors contributed equally.\par
      \textsuperscript{\#}Corresponding authors: Qian Niu.\par
      Correspondence to: \icmlcorrespondingauthor@text.\par}
  \fi}
\newcommand{\aivcprintnotice}{%
  \ificmlshowauthors
    \global\icml@noticeprintedtrue
    {\let\thefootnote\relax\footnotetext{\hspace*{-\footnotesep}\Notice@String}}%
  \else
    \printAffiliationsAndNotice{}%
  \fi}
\makeatother

\icmltitlerunning{scAnyOmics: Foundation-Model Distillation for Any-to-Any Completion of Multi-omics Virtual Cells}

\begin{document}

\twocolumn[
  \icmltitle{scAnyOmics: Foundation-Model Distillation for Any-to-Any Completion \\
    of Multi-omics Virtual Cells}

  \icmlsetsymbol{equal}{*}
  \icmlsetsymbol{corr}{\#}

  \begin{icmlauthorlist}
    \icmlauthor{Zhaoqi Song}{utokyo,xmu,equal}
    \icmlauthor{Ye Du}{polyu,equal}
    \icmlauthor{Yusuke Iwasawa}{utokyo}
    \icmlauthor{Yutaka Matsuo}{utokyo}
    \icmlauthor{Qian Niu}{utokyo,corr}
  \end{icmlauthorlist}

  \icmlaffiliation{utokyo}{Department of Technology Management for Innovation, Graduate School of Engineering,\\ The University of Tokyo, Tokyo, 113-8656, Japan}
  \icmlaffiliation{xmu}{State Key Laboratory of Cellular Stress Biology, School of Life Sciences, Faculty of Medicine and Life Sciences, \\ Xiamen University, Xiamen, 361102, China}
  \icmlaffiliation{polyu}{Department of Computing, Hong Kong Polytechnic University, Hong Kong S.A.R., China}

  \icmlcorrespondingauthor{Qian Niu}{qian.niu@weblab.t.u-tokyo.ac.jp}
  \aivcaffilblock

  \icmlkeywords{virtual cell, single-cell multi-omics, knowledge distillation, foundation models, missing-modality inference}

  \vskip 0.22in
]

\aivcprintnotice

\begin{abstract}
Recent advances in single-cell technologies have enabled the profiling of multiple molecular modalities, including transcriptomics, proteomics, and chromatin accessibility, providing a more comprehensive understanding of cellular states.
However, integrating heterogeneous modalities and predicting missing molecular information remain challenging, particularly when paired measurements are limited.
Here, we introduce \textbf{scAnyOmics}, a generative framework that distills knowledge from a frozen single-cell foundation model into a multimodal student model. scAnyOmics integrates scRNA-seq, protein-marker measurements, and scATAC-seq into a shared representation, enabling cross-modal prediction and reconstruction of missing modalities.
We assembled 690,880 cells from six publicly available studies, covering both paired and partially paired multi-omics measurements, to systematically evaluate its predictive performance.
We further evaluated scAnyOmics on held-out donor data, where it outperformed baseline methods across multiple evaluation metrics.
Beyond cross-modal reconstruction, we explored its applications in cell-type annotation, perturbation-response prediction, and biological age estimation.
Overall, scAnyOmics provides a unified framework for integrating incomplete single-cell multi-omics data and transferring pretrained biological knowledge, supporting the development of virtual cell models across diverse cellular contexts and experimental platforms.
\end{abstract}

\section{Introduction}
\label{sec:intro}

\begin{figure*}[t]
  \centering
  \includegraphics[width=\textwidth]{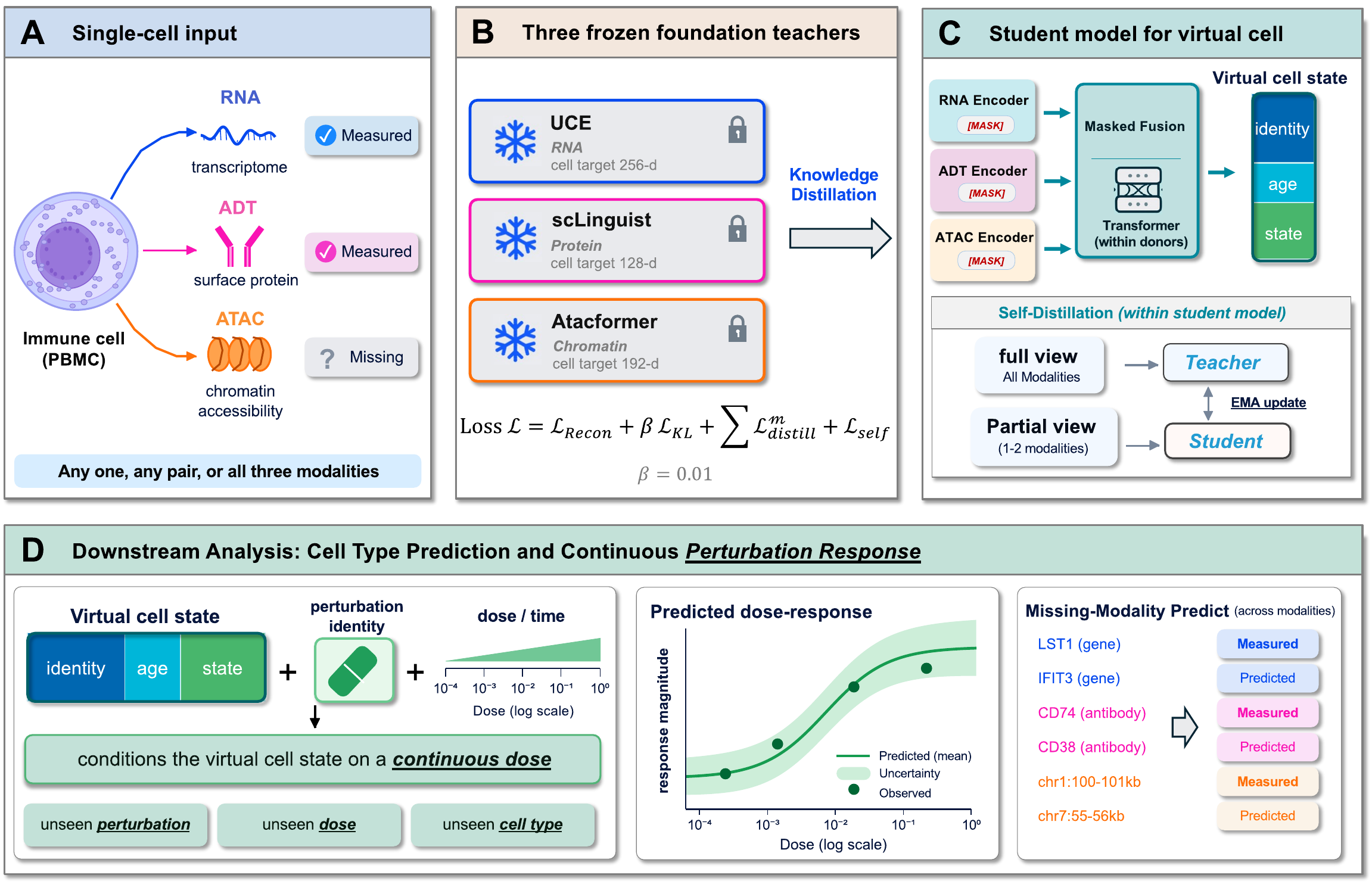}
  \caption{\textbf{Overview of the scAnyOmics framework.} scAnyOmics contains four main components: \textbf{(A)} single-cell multi-omics input, including scRNA-seq, surface proteins (ADT), and chromatin accessibility (ATAC); \textbf{(B)} knowledge distillation from three frozen foundation models (\textbf{UCE}, \textbf{scLinguist}, and \textbf{Atacformer}); \textbf{(C)} multimodal student training with masked inputs and self-distillation to learn unified cellular representations; and \textbf{(D)} downstream applications, including cell-type prediction, continuous perturbation-response prediction, and missing-modality inference.}
  \label{fig:overview}
\end{figure*}

Cells are complex biological systems whose states are shaped by interactions among multiple molecular layers.
Single-cell sequencing technologies now allow researchers to examine these layers, including gene expression, protein markers, and chromatin accessibility, at unprecedented resolution.
Each modality provides a different perspective on cellular identity and function.
Gene expression reflects transcriptional activity, chromatin accessibility provides information about gene regulation, and protein measurements reveal important functional characteristics of cells.
Integrating these measurements can therefore provide a more comprehensive understanding of cellular states.
However, most single-cell datasets contain only a subset of these modalities, making it difficult to build a unified representation of cells and predict molecular information that has not been directly measured.

The growing interest in virtual cell models offers a new direction for addressing this challenge.
Virtual cells aim to represent biological systems computationally and predict cellular properties or responses under different conditions~\citep{bunne2024virtual}.
Recent studies, such as State and ProteinTalks, have demonstrated how computational models can predict cellular responses to perturbations and capture changes in molecular states~\citep{adduri2026state,sun2026proteintalks}.
These advances highlight the potential of computational models to support biological discovery and \emph{in silico} experiments.
An important step toward this goal is to develop models that can integrate different molecular measurements and reconstruct missing modalities.
Such capabilities would allow researchers to study cellular states more comprehensively, even when only partial molecular information is available.

Recent advances in single-cell multi-omics technologies have created opportunities to develop these models.
CITE-seq enables the simultaneous measurement of gene expression and selected surface proteins~\citep{stoeckius2017citeseq}, while TEA-seq and DOGMA-seq further combine transcriptomic, chromatin accessibility, and protein-marker measurements within individual cells~\citep{swanson2021teaseq,mimitou2021dogmaseq}.
However, generating fully paired multi-omics datasets remains technically demanding, and many existing datasets contain only partially overlapping modalities.
Differences in measurement platforms and protein panels also make it difficult to combine data across studies.
Consequently, computational models must be able to learn from both paired and unpaired measurements while distinguishing genuinely missing information from observed values.

Several computational methods have been developed to address multimodal integration and missing-data prediction.
For example, totalVI jointly models RNA and protein measurements while accounting for technical variation~\citep{gayoso2021totalvi}.
MultiVI extends this approach to paired and unpaired multimodal datasets and enables missing-modality inference~\citep{ashuach2023multivi}.
Similarly, scVAEIT incorporates missingness information to integrate gene expression, protein abundance, and chromatin accessibility across heterogeneous datasets~\citep{du2022scvaeit}.
These methods have established important foundations for single-cell multi-omics integration.
Nevertheless, a key opportunity remains to investigate whether knowledge learned from large-scale pretrained models can further improve cross-modal prediction, particularly when training data are incomplete or vary across experimental platforms.

Single-cell foundation models provide a promising way to incorporate such knowledge.
Models including Geneformer and scGPT have shown that representations learned from large single-cell datasets can be transferred to downstream biological tasks~\citep{theodoris2023geneformer,cui2024scgpt}.
Universal Cell Embedding (UCE) further demonstrates how gene-expression profiles from different tissues and species can be mapped into a shared representation space~\citep{rosen2026uce}.
More recently, HoloCell has explored multimodal foundation modelling for integrating gene expression, chromatin accessibility, and protein markers, including the generation of missing modalities~\citep{jiang2026holocell}.
Together, these developments suggest that pretrained representations can provide useful biological information for multimodal learning.
However, effectively transferring this information into a specialized model remains an important challenge, particularly when the goal is to make accurate predictions from heterogeneous and partially observed measurements.

Knowledge distillation offers a practical approach to this challenge.
Originally developed to transfer information from a pretrained teacher model to a student model~\citep{hinton2015distilling}, knowledge distillation allows the student to learn from rich representations without requiring the teacher during inference.
In single-cell multi-omics analysis, this approach could enable a student model to combine information from different molecular modalities while benefiting from the biological knowledge encoded by a foundation model.
By keeping the teacher frozen and training a separate multimodal student, it becomes possible to investigate whether pretrained knowledge improves the reconstruction of missing molecular measurements.
This strategy also creates an opportunity to distinguish the contribution of foundation-model representations from that of the student's own multimodal learning.

Evaluating these models requires careful attention to prediction accuracy and generalization.
Previous studies have shown that single-cell foundation models do not consistently outperform simpler approaches across all datasets and tasks~\citep{kedzierska2025zeroshot}.
Similarly, recent perturbation-prediction benchmarks have demonstrated the importance of selecting appropriate baselines and evaluation metrics~\citep{ahlmanneltze2025perturbation,miller2026calibrated}.
Single-cell datasets also contain biological and technical dependencies, such as multiple cells originating from the same donor, which can affect performance estimates if they are not considered during evaluation~\citep{zimmerman2021pseudoreplication}.
Therefore, reliable assessment of multimodal prediction should include comparisons with established methods, evaluation across different molecular targets, and testing on independent biological samples or datasets.

Here, we introduce \textbf{scAnyOmics}, a multi-teacher knowledge distillation framework for integrating single-cell multi-omics data and predicting missing molecular modalities. scAnyOmics leverages three frozen foundation models---UCE for gene expression, scLinguist for surface proteins, and Atacformer for chromatin accessibility---to transfer modality-specific knowledge into a unified multimodal student model.
Through knowledge distillation, masked multimodal learning, and self-distillation, scAnyOmics learns shared cellular representations from incomplete molecular measurements, enabling cross-modal prediction and reconstruction.
We assembled \textbf{690,880 cells from six publicly available studies}, covering paired and partially paired measurements of gene expression, protein markers, and chromatin accessibility.
Using these datasets, we systematically evaluate cross-modal prediction and investigate how pretrained representations and different learning strategies contribute to model performance.
We further assess the quality of learned cellular representations through linear and nonlinear prediction probes and evaluate model generalization across independent samples.
Overall, scAnyOmics provides a unified framework for transferring knowledge from multiple foundation models into multimodal cellular representations, supporting the reconstruction of missing molecular information and the development of computational virtual cell models.

\section{Results}
\label{sec:results}

\begin{figure*}[t]
  \centering
  \includegraphics[width=\textwidth]{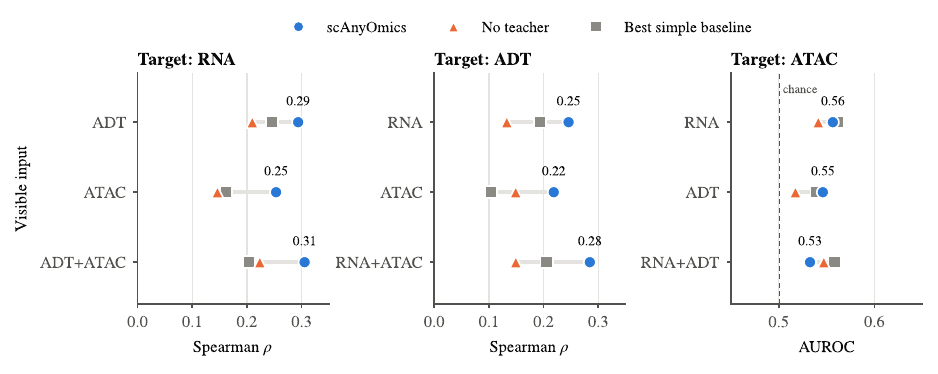}
  \caption{\textbf{Reconstruction performance across nine input--target combinations.}
    Each panel corresponds to a withheld target modality, with rows indicating the available input modality or modalities.
    \textbf{Blue} circles denote the UCE-distilled scAnyOmics student, \textbf{orange} triangles the student without teacher distillation, and \textbf{grey} squares the stronger of ridge regression and 15-nearest-neighbour averaging.
    RNA and ADT outcomes are median Spearman correlations across 2,000 genes and 166 antibodies, respectively; ATAC outcomes are median AUROC values across 8,000 10-kb genomic bins.
    The vertical dashed line indicates chance-level AUROC (0.50).
    Numbers beside the blue markers are scAnyOmics scores.
    All tasks were evaluated on the same 28,239 trimodal cells from held-out samples.}
  \label{fig:completion}
\end{figure*}

\begin{figure*}[t]
  \centering
  \includegraphics[width=\textwidth]{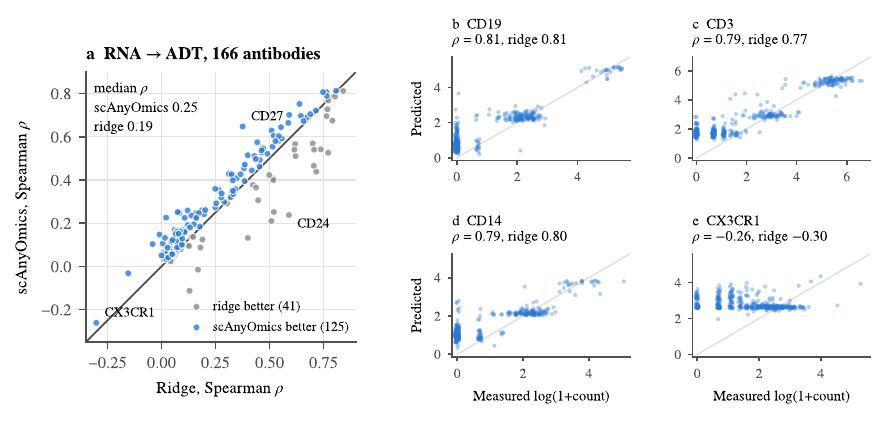}
  \caption{\textbf{Antibody-specific accuracy of RNA-to-ADT reconstruction.}
    \textbf{(a)} Spearman correlation for each of 166 antibodies predicted from RNA, comparing ridge regression (horizontal axis) with scAnyOmics (vertical axis), using 10,000 trimodal test cells.
    Markers above the diagonal favour scAnyOmics (125 antibodies); markers below it favour ridge regression (41 antibodies).
    \textbf{(b--e)} Measured and predicted $\log(1 + \mathrm{count})$ values for CD19, CD3, CD14, and CX3CR1 in a random subset of 400 test cells.
    Panel annotations report the Spearman correlations of scAnyOmics and ridge regression.
    Measured values were jittered horizontally for visualization.}
  \label{fig:per_antibody}
\end{figure*}

\begin{figure*}[t]
  \centering
  \includegraphics[width=\textwidth]{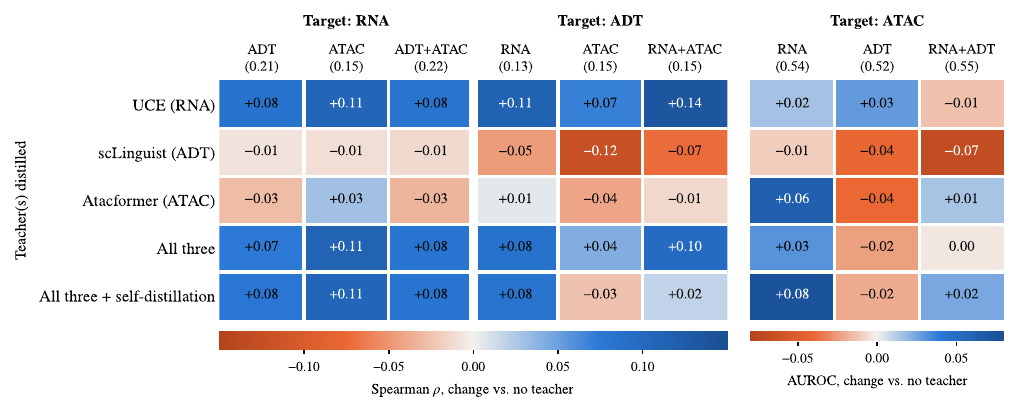}
  \caption{\textbf{Effects of foundation-model teachers and self-distillation on cross-modal reconstruction.}
    Rows indicate the teacher configuration and columns the observed input--target task.
    Entries show differences relative to a student trained without teacher distillation; positive changes (blue) indicate higher reconstruction accuracy and negative changes (orange) indicate lower accuracy.
    RNA and ADT panels report changes in median Spearman correlation, whereas the ATAC panel reports changes in median AUROC on a separate colour scale.
    Values in parentheses beneath each task label give the corresponding teacher-free score.
    All comparisons use the same 28,239 trimodal held-out sample cells as \cref{fig:completion}.}
  \label{fig:teachers}
\end{figure*}

\subsection{Cross-modal reconstruction from incomplete multi-omics profiles}

We evaluated the scAnyOmics ability to reconstruct unseen modalities from one or two available modalities.
Using 28,239 cells with matched RNA, antibody-derived tag (ADT), and chromatin-accessibility (ATAC) measurements in held-out samples, we masked the target modality and assessed all nine input--target combinations \textbf{(\cref{fig:completion})}.
Reconstruction accuracy was summarized by the median Spearman correlation across 2,000 genes or 166 antibodies for RNA and ADT, respectively, and the median area under the receiver operating characteristic curve (AUROC) across 8,000 genomic bins for ATAC.
We compared the UCE-distilled student, trained on 482,102 cells from the original four-study assembly, with an otherwise comparable student without teacher distillation and with the better of ridge regression and 15-nearest-neighbour prediction for each task.

The UCE-distilled model outperformed the simple baselines in all six tasks with RNA or ADT as the reconstruction target (\cref{fig:completion}).
Across these tasks, the improvement in median Spearman correlation was approximately 0.05--0.12.
The largest difference occurred when ATAC was the sole input for ADT prediction ($\rho = 0.22$ versus 0.10 for the best simple baseline).
ATAC-to-RNA prediction also improved ($\rho = 0.25$ versus 0.16).
Combining two observed modalities further improved RNA and ADT reconstruction in the evaluated settings: ADT prediction increased from 0.25 with RNA alone to 0.28 with RNA and ATAC, while RNA prediction reached 0.31 with ADT and ATAC, compared with 0.29 using ADT alone.
The teacher-free student scored 0.07--0.14 lower than the UCE-distilled model across all six RNA and ADT tasks and was outperformed by the simple baselines in four of them.

For ATAC prediction, scAnyOmics achieved above-chance performance across different input modalities, with median AUROC values ranging from 0.53 to 0.56 (\cref{fig:completion}).
The highest performance was obtained using RNA alone (AUROC = 0.56), followed by ADT alone (AUROC = 0.55), indicating that both transcriptomic and protein measurements contain information relevant to chromatin accessibility.
Combining RNA and ADT yielded an AUROC of 0.53.
Although performance remained comparable to the best simple baselines, these results suggest that scAnyOmics can capture cross-modal associations between molecular measurements and chromatin accessibility, providing a basis for further improving ATAC reconstruction.

\subsection{Antibody-level variation in RNA-to-protein prediction}

We next examined how RNA-to-ADT prediction varied across individual protein markers.
In an analysis of 10,000 trimodal test cells, scAnyOmics achieved a higher Spearman correlation than ridge regression for 125 of the 166 antibodies (75.3\%; \cref{fig:per_antibody}a).
The median correlation across antibodies was 0.25 for scAnyOmics and 0.19 for ridge regression.
The improvement was concentrated among markers with intermediate prediction accuracy, whereas several well-predicted lineage-associated markers showed similar performance under both methods.
For CD19, the correlations were 0.81 for both models; for CD3, they were 0.79 and 0.77; and for CD14, 0.79 and 0.80, respectively.
Ridge regression retained an advantage for 41 antibodies, including CD24.
Both methods performed poorly for CX3CR1 ($\rho = -0.26$ for scAnyOmics and $-0.30$ for ridge regression).

The measured-versus-predicted profiles further characterized the resolution of RNA-based protein reconstruction (\cref{fig:per_antibody}b--e).
For CD19, CD3, and CD14, predicted expression separated groups of cells with different measured marker levels, consistent with the high correlations observed for these antibodies.
Within those groups, however, predictions occupied narrower ranges than the measurements.
This pattern indicates that performance for these markers was driven largely by differences between cellular populations, while variation among cells with similar marker expression profiles was recovered less accurately.
CX3CR1 showed weak correspondence between predicted and measured values, consistent with its negative rank correlation.
These results demonstrate substantial antibody-to-antibody heterogeneity in reconstruction accuracy and distinguish recovery of broad population-level patterns from prediction of finer within-population variation.

\subsection{Effects of modality-specific teacher models and self-distillation}

To investigate how different foundation models contribute to cross-modal reconstruction, we compared scAnyOmics trained with each teacher separately (UCE, scLinguist, and Atacformer), with all three teachers combined, and with additional self-distillation (\cref{fig:teachers}).
We used a student model without knowledge distillation as the baseline.
\Cref{fig:teachers} shows the improvement in prediction performance for each configuration, measured by changes in median Spearman correlation for RNA and ADT and median AUROC for ATAC.

Among the three foundation models, \textbf{UCE provided the most consistent improvements} across prediction tasks.
Using UCE alone improved all six RNA and ADT reconstruction tasks, with gains of 0.07--0.14 in Spearman correlation.
The largest improvement was observed when predicting ADT from RNA and ATAC together (+0.14).
UCE also improved RNA prediction from ATAC alone (+0.11) and ADT prediction from RNA alone (+0.11).
These results highlight the value of pretrained transcriptomic representations for predicting molecular information across different modalities.

Combining all three teachers also improved RNA and ADT reconstruction across all six tasks.
The gains ranged from 0.07 to 0.11 for RNA prediction and from 0.04 to 0.10 for ADT prediction.
Atacformer provided an additional advantage for chromatin accessibility, improving RNA-to-ATAC prediction by 0.06 AUROC when used alone.
\textbf{Self-distillation further improved RNA-to-ATAC prediction}, increasing the gain from +0.03 with all three teachers to +0.08, corresponding to an AUROC of 0.62 compared with 0.54 for the teacher-free baseline.
The self-distillation model also maintained improvements of 0.08--0.11 across all three RNA reconstruction tasks.

The effects of knowledge distillation varied across modalities.
While UCE consistently improved RNA and ADT predictions, scLinguist alone did not improve reconstruction performance in the evaluated tasks.
Adding self-distillation also reduced performance in some ADT prediction settings.
Overall, these results demonstrate that foundation-model knowledge can improve cross-modal reconstruction, with UCE contributing the most consistent benefits and self-distillation providing further gains in chromatin accessibility prediction.

\section{Methods}
\label{sec:methods}

\subsection{Study design and prediction task}

We designed scAnyOmics as a virtual cell framework to integrate single-cell multi-omics data, including gene expression (RNA), surface proteins (ADT), and chromatin accessibility (ATAC), as presented in \textbf{\cref{fig:overview}A}. scAnyOmics learns a shared representation of cellular states from these modalities, enabling the prediction of missing molecular measurements when only partial information is available.

For each cell, we used 2,000 genes, 166 protein markers, and 8,000 chromatin-accessibility bins.
Since different datasets contain different combinations of modalities, we used measurement masks to distinguish available data from missing modality.
During training, measured modalities could also be hidden from the model and used as prediction targets.
For protein measurements, we further recorded which proteins were measured in each cell, ensuring that unmeasured proteins were excluded from training and evaluation.

scAnyOmics uses separate encoders to process each modality and combines their information into a shared cellular representation \textbf{(\cref{fig:overview}C)}.
A Transformer incorporates information from other cells within the same sample group, while knowledge distillation transfers pretrained representations from respective frozen foundation model to the student model, in which we choose UCE for RNA, scLinguist for ADT and Atacformer for ATAC \textbf{(\cref{fig:overview}B)}.
Separate decoders then reconstruct the molecular measurements for each modality.
Once trained, the student model can infer missing modalities without requiring the teacher-foundation model.
After knowledge-distillation training, scAnyOmics also aims to resolve the downstream tasks including cell-type annotation and perturbation response \textbf{(\cref{fig:overview}D)}.

\subsection{Datasets}

We assembled 690,880 cells from six Gene Expression Omnibus (GEO) series profiling human peripheral blood \textbf{(\cref{tab:datasets})}: GSE158013 (59,151 cells), GSE200417 (68,138), GSE214546 (367,955), GSE236422 (181,518), GSE156478 (5,527) and GSE166188 (8,591).
Four series (GSE158013, GSE214546, GSE236422 and GSE166188) profiled peripheral blood mononuclear cells (PBMCs) or blood-derived T cells with trimodal assays (TEA-seq or DOGMA-seq).
GSE200417 compared DOGMA-seq with CITE-seq in activated pan-T cells, and GSE156478 contributed one CITE-seq library.
The cohorts differ in biological state.
GSE158013 and GSE214546 profiled unstimulated PBMCs.
GSE200417 profiled T cells stimulated for 12\,h with CD3/CD28/CD2 activator, IL-1$\beta$ and IL-23, with PGE2 and/or TGF-$\beta$.
GSE236422 profiled healthy-donor T cells treated \emph{in vitro} with bortezomib, lenalidomide, dexamethasone or DMSO for 4, 24 or 72\,h, plus an untreated 0\,h sample.
The GSE156478 and GSE166188 libraries were the control (16\,h IL-2) arms of a CD3/CD28 stimulation experiment.

The data were organized into 47 sample-directory blocks, which correspond to source libraries or processing units, not to confirmed donors.
The 47 blocks derive from 44 GEO sample (GSM) accessions; for three GSE200417 DOGMA-seq libraries, cells lacking an ADT barcode match were stored as separate RNA--accessibility blocks.
Donor identity was available from GEO for only two series.
All five GSE158013 TEA-seq libraries and both 10x Multiome libraries came from a single donor (HMN200910).
The 19 GSE214546 libraries came from 16 donors and one control individual.
The donors comprised eight adults aged 55--63 years and eight 11-year-old children, all female; the control individual (IMM19, male, age not reported) contributed three libraries.
Donor identity could not be resolved for the other four series.

The combined matrix contained 309,724 cells with all three modalities, 353,018 RNA--ADT cells, 26,094 RNA--accessibility cells and 2,044 RNA-only cells \textbf{(\cref{tab:cells})}.
The RNA-only cells were GSE158013 10x Multiome cells whose barcodes could not be matched to fragment records.
RNA--ADT cells arose in two ways.
Some came from CITE-seq libraries (GSE200417 GSM6032900 and GSE156478).
The others were trimodal-assay cells without matched fragments.
These included all cells from the GSE214546 B069 and B076 sequencing batches (14 libraries), whose fragment-file barcodes did not match the expression-matrix barcodes, and 3--13\% of cells in the remaining libraries.
Thus, only the five B065-batch libraries of GSE214546 contributed accessibility data.
The assembly extended an earlier 633,803-cell, four-series dataset (GSE158013 TEA-seq, GSE200417 DOGMA-seq, GSE214546 and GSE236422) with 57,077 cells from eight additional blocks.
The rows of the earlier dataset were left unchanged, and all added cells were assigned to training.
We used the cells retained in each source's filtered count matrices.
Filtering criteria and sample annotations therefore followed each source study and were not uniform across studies.

\begin{table*}[t]
  \caption{\textbf{Source datasets.}}
  \label{tab:datasets}
  \centering
  \footnotesize
  \setlength{\tabcolsep}{4pt}
  \renewcommand{\arraystretch}{1.15}
  \begin{tabularx}{\textwidth}{@{}>{\raggedright\arraybackslash}p{1.75cm}
                    >{\raggedright\arraybackslash}X
                    >{\raggedright\arraybackslash}X
                    >{\raggedright\arraybackslash}X
                    c
                    >{\raggedright\arraybackslash}p{1.9cm}
                    c@{}}
    \toprule
    \textbf{GEO series (PMID)} & \textbf{Biological material} & \textbf{Condition} & \textbf{Assay(s)} & \textbf{\shortstack{Blocks /\\GSMs}} & \textbf{Donors} & \textbf{\shortstack{Anti-\\bodies}} \\
    \midrule
    GSE158013 (33835024) & Leukapheresis PBMCs, dead-cell/neutrophil-depleted & Unstimulated & TEA-seq (5 libraries); 10x Multiome, nuclei or permeabilized cells (2 libraries) & 7 / 7 & 1 (HMN200910) & 45 \\
    GSE200417 (35739535) & Peripheral blood pan-T cells, healthy donors aged 18--35 years & 12\,h CD3/CD28/CD2 + IL-1$\beta$ + IL-23 $\pm$ PGE2/TGF-$\beta$ & DOGMA-seq, DIG (2 libraries) and LLL (1 library) permeabilization; CITE-seq (1 library) & 7 / 4 & Not resolved & 154 \\
    GSE214546 (37845489) & PBMCs & Unstimulated & TEA-seq & 19 / 19 & 16 donors + 1 control individual & 54 \\
    GSE236422 (---) & Blood T cells from healthy donors & Bortezomib, lenalidomide, dexamethasone or DMSO $\times$ 4/24/72\,h; untreated 0\,h & TEA-seq & 12 / 12 & Not resolved & 54 \\
    GSE156478 (34083792) & PBMCs & Control arm (16\,h IL-2) & CITE-seq & 1 / 1 & Not resolved & 113 \\
    GSE166188 (34083792) & PBMCs & Control arm (16\,h IL-2) & DOGMA-seq, LLL permeabilization & 1 / 1 & Not resolved & 107 \\
    \bottomrule
  \end{tabularx}
\end{table*}

\begin{table*}[t]
  \caption{\textbf{Cells per dataset and modality combination.}}
  \label{tab:cells}
  \centering
  \small
  \begin{tabular}{@{}lrrrrrr@{}}
    \toprule
    \textbf{GEO} & \textbf{Total cells} & \textbf{RNA+ADT+ATAC} & \textbf{RNA+ADT} & \textbf{RNA+ATAC} & \textbf{RNA} & \textbf{Split (train / test)} \\
    \midrule
    GSE158013 & 59,151  & 37,745  & 4,682   & 14,680 & 2,044 & 49,911 / 9,240 \\
    GSE200417 & 68,138  & 41,903  & 14,821  & 11,414 & 0     & 56,195 / 11,943 \\
    GSE214546 & 367,955 & 53,423  & 314,532 & 0      & 0     & 246,439 / 121,516 \\
    GSE236422 & 181,518 & 168,062 & 13,456  & 0      & 0     & 172,516 / 9,002 \\
    GSE156478 & 5,527   & 0       & 5,527   & 0      & 0     & 5,527 / 0 \\
    GSE166188 & 8,591   & 8,591   & 0       & 0      & 0     & 8,591 / 0 \\
    \midrule
    \textbf{Total} & \textbf{690,880} & \textbf{309,724} & \textbf{353,018} & \textbf{26,094} & \textbf{2,044} & \textbf{539,179 / 151,701} \\
    \bottomrule
  \end{tabular}
\end{table*}

\subsection{Data preprocessing and feature harmonization}

We harmonized the source-specific count matrices, barcodes, fragment files and peak files into a common feature space.
Features were selected only on the trimodal training cells of the four-series assembly (historical split), so that no test cells influenced feature selection.

\paragraph{RNA.}
The four original series shared a 36,601-gene reference.
For each gene $g$ we computed the variance of log-transformed, unnormalized counts across the feature-selection set $H$:
\begin{align}
  u_{ig} &= \log\bigl(1 + x_{ig}^{R}\bigr), \\
  V_{g} &= \max\Biggl\{ \frac{1}{n_{H}}\sum_{i \in H}u_{ig}^{2} - \Bigl( \frac{1}{n_{H}}\sum_{i \in H}u_{ig} \Bigr)^{2},\ 0 \Biggr\},
\end{align}
where $H$ is the set of trimodal training cells and $n_{H} = |H|$.
We retained the 2,000 genes with the largest $V_{g}$, in their original column order; these highly variable genes (HVGs) were therefore selected by empirical log-count variance.
Datasets added later were mapped to this fixed panel by gene symbol rather than column position, because GSE156478 used a different (33,538-gene) reference.
When a symbol was duplicated, the first occurrence was kept; genes absent from a source were filled with zero-valued columns.
GSE156478 lacked 33 of the 2,000 selected genes, and no other source lacked any.
The gene panel was not recalculated for later validation splits.

\paragraph{ADT.}
Antibody names were canonicalized by removing terminal TotalSeq-A clone suffixes, standardizing capitalization and punctuation, and applying receptor-name aliases (for example, HLA.DR/HLADR).
Isotype-control features matching \texttt{IGG.*(CTRL\textbar CONTROL)} were removed.
When several source features mapped to the same canonical name, the first was retained.
The analysis used a fixed union panel of 166 antibodies defined on the four original series.
The 116 additional antibodies introduced by the added sources were excluded, not masked; consequently, GSE156478 and GSE166188 contributed 113 and 107 of their measured antibodies, respectively (\cref{tab:datasets}).
A cell-by-antibody mask recorded which targets were measured in each cell, and unmeasured antibodies were excluded from training losses and evaluation metrics rather than treated as zero.

\paragraph{Chromatin accessibility (ATAC).}
We represented accessibility with 10-kb bins on hg38 chromosomes 1--22 and X.
Each fragment was assigned to the bin containing its midpoint, and each fragment-file row was counted once, ignoring the optional duplicate-count column.
For bin $k$ we defined $b_{ik} = \mathbf{1}[c_{ik} > 0]$ and $p_{k} = n_{H}^{-1}\sum_{i \in H}b_{ik}$, and retained the 8,000 bins with the largest Bernoulli variance $p_{k}(1 - p_{k})$.
Accessibility values were then binarized.
GSE166188 provided only a called-peak matrix and no fragment file, so peak counts were assigned to the bin containing each peak midpoint before binarization.
Cells without matched fragments received an empty accessibility row with the accessibility mask set to zero; this row was excluded from all losses, teacher targets and evaluations.
These features describe accessibility at 10-kb resolution, not individual regulatory elements.

\subsection{Sample source and training-evaluation splits}
\label{sec:splits}

We first divided the 690,880 cells into 539,179 cells for training set and 151,701 cells for test set.
In datasets with multiple blocks, we reserved at least one block and approximately 20\% of cells for testing.
The 57,077 cells added during dataset expansion entered the original training set.

We checked sample and library identifiers against the available source metadata.
We could identify one donor in GSE158013 and 17 individuals in GSE214546, but could not determine donor identities for the other four datasets.
Donor HMN200910 appeared in both training and test libraries.
Source library GSM6032893 also contributed cells to a test block and an additional training block.
We recorded these overlaps and tested a stricter grouping approach, assigning samples with unknown donor identity to their source dataset.
That approach left no trimodal test cells, so the following reconstruction analyses use the original sample-held-out cohorts.

For RNA-to-ADT prediction, we selected cells with measured RNA and ADT and an available UCE embedding.
We reserved one historical training sample from each represented dataset for validation (sample indices 3, 7, 26, and 38).
Using random seed 17, we selected 60,000 cells for probe training, 6,000 for validation, 8,000 from the RNA--ADT test cohort, and 8,000 from the trimodal test cohort.
Sampling was without replacement within each cohort; the two test cohorts may overlap.
All 166 antibodies were measured in at least 30 cells in each cohort and were eligible for evaluation.
We selected them using measurement masks, without examining test values.

For the B, B-norm, and B-tech comparisons, we removed training blocks that shared a source-library accession with any validation or test block, including blocks without ADT.
This left 477,439 training cells.
All three configurations used the same validation and test cohorts.
This exclusion removed shared-library overlap, but not the confirmed overlap of donor HMN200910 across separate libraries.
Feature selection and UCE preprocessing had also taken place before this revised split.
We consequently treat the results as sample-held-out prediction tests, rather than independent-donor generalization tests.

\subsection{Frozen foundation-model teachers and distillation targets}

scAnyOmics is designed with three frozen foundation-model teachers, one for each measured modality: UCE for RNA, scLinguist for ADT, and Atacformer for ATAC.
Their representations provide modality-specific learning targets for a shared student model.
\Cref{fig:overview} shows this three-teacher architecture.
The B, B-norm, and B-tech experiments reported here used the UCE branch; scLinguist and Atacformer were not included in these training runs.
For the UCE branch, the teacher produced a 1,280-dimensional embedding for each cached cell.
We used principal component analysis (PCA), fitted on historical training embeddings, to reduce it to 256 dimensions.
Let $e_{i} \in \mathbb{R}^{1280}$ denote an original UCE embedding, $\bar{e}_{H}$ the historical training mean, and $V_{256} \in \mathbb{R}^{256 \times 1280}$ the PCA loading matrix.
We calculated the distillation target as follows:
\begin{align}
  u_{i} &= V_{256}\bigl( e_{i} - \bar{e}_{H} \bigr), \qquad
  \bar{u}_{H} = \frac{1}{|H_{T}|}\sum_{i \in H_{T}}u_{i}, \\
  s_{H} &= \sqrt{\frac{1}{256}\sum_{d = 1}^{256}\Var_{i \in H_{T}}\bigl( u_{id} \bigr)}, \\
  t_{i} &= \frac{u_{i} - \bar{u}_{H}}{\max(s_{H},10^{-8})}.
\end{align}
Here $H_{T}$ denotes the cells used to build the original teacher cache.
A single scale factor $s_{H}$ standardizes the projected embeddings while preserving differences in variance across PCA coordinates.
For sample group $g$, we also averaged its cached cell embeddings to create a group-level target:
\begin{equation}
  c_{g} = \frac{1}{|C_{g}|}\sum_{i \in C_{g}}t_{i}.
\end{equation}
Here $C_{g}$ denotes the cached cells in sample group $g$.
This average summarizes the UCE representations of that sample group.

The UCE cache covered 633,802 of 690,880 cells.
The remaining 57,078 cells, including all 57,077 newly added cells, had no UCE distillation target in the reported runs.
We kept the PCA projection and scaling fitted during the original cache preparation.
Embeddings computed after data expansion were not used for B, B-norm, or B-tech.
The same UCE preprocessing preceded the revised validation split.

The three-teacher design assigns separate distillation targets to the RNA, ADT, and ATAC branches.
The architecture specifies 256-dimensional UCE targets, 128-dimensional scLinguist targets, and 192-dimensional Atacformer targets.
In the complete design, each teacher remains frozen, and the student would match the corresponding target when that modality has been measured.
The present experimental results describe the UCE-only implementation; the other two teacher branches require separate training and evaluation before their effects can be reported.

\subsection{Modality encoders and fusion}

In the B and B-tech models, each modality encoder received log-transformed measurements together with a log-transformed total count.
For modality $m$, the input was:
\begin{equation}
  f_{i}^{m} = \bigl[ \log(1 + x_{i}^{m});\, q_{i}^{m} \bigr], \quad
  q_{i}^{m} = \log\Bigl( 1 + \sum_{j}x_{ij}^{m} \Bigr).
\end{equation}
We set negative count values to zero before transformation.
ATAC inputs were binary.
In the baseline ADT encoder, missing antibody entries were stored as zeros, and the antibody mask was not part of the encoder input.
The mask was still used to ensure that only measured antibodies contributed to training losses and evaluation, as indicated by $a_{ij}$.

Each modality was processed by a separate two-layer neural network.
Both layers had 512 units; the network used layer normalization, GELU activation, and dropout of 0.1:
\begin{equation}
  \begin{split}
    h_{i}^{m} = \GELU\Bigl( &W_{2}^{m}\Drop_{0.1}[\GELU(\LN(W_{1}^{m}f_{i}^{m} + b_{1}^{m}))] \\
    &+ b_{2}^{m} \Bigr).
  \end{split}
\end{equation}
Each encoder produced a mean $\mu_{i}^{m}$ and log variance $\ell_{i}^{m}$ for the shared representation, a mean $\mu_{i}^{m,p}$ and log variance $\ell_{i}^{m,p}$ for a modality-specific representation, and a reliability score:
\begin{equation}
  r_{i}^{m} = 0.1 + 0.9\,\sigma\bigl(w_{r}^{m\top}h_{i}^{m} + b_{r}^{m}\bigr).
\end{equation}
Here $\sigma$ is the sigmoid function.
We limited shared log variances to $[-8,2]$ and private log variances to $[-8,6]$.
The shared representation had 96 dimensions.
The code grouped these into 32 identity, 16 age, and 48 state coordinates.
These are names for blocks of coordinates; the reported training objective did not separately train them to predict identity, age, or cell state.

We combined information from the visible modalities using a reliability-weighted product of Gaussian experts.
This gives more weight to modalities with higher estimated precision and also includes a standard-normal prior.
For coordinate $d$:
\begin{align}
  \tau_{id}^{m} &= v_{i}^{m}r_{i}^{m}\exp(-\ell_{id}^{m}), \qquad
  T_{id} = 1 + \sum_{m \in \mathcal{M}}\tau_{id}^{m}, \\
  \mu_{id}^{(0)} &= \frac{\sum_{m}\tau_{id}^{m}\mu_{id}^{m}}{T_{id}}, \qquad
  \ell_{id}^{F} = -\log T_{id}.
\end{align}
The prior has mean zero and precision one.
A modality contributes more strongly when its reliability score is higher or its estimated variance is lower.
The resulting fusion weights are:
\begin{equation}
  w_{id}^{m} = \tau_{id}^{m}/T_{id}, \qquad w_{id}^{\mathrm{prior}} = 1/T_{id}.
\end{equation}
These weights determine how the encoder combines modalities.
We did not interpret them as direct measurements of biological importance.

\subsection{Context from cells in the same patient sample}

We processed cells in groups of up to 256 drawn from the same sample.
A one-layer Transformer allowed each cell to use information from the other cells in its group.
The Transformer had four attention heads of width 24, a feed-forward layer of width 512, GELU activation, and dropout of 0.1.
It took the matrix of fused cell representations $X$ as input, without positional embeddings.
Attention was calculated as:
\begin{equation}
  \mathrm{Attn}_{\mathrm{train}}(Q,K,V) = \Drop_{0.1}\Bigl[ \softmax_{\mathrm{row}}\Bigl( \frac{QK^{\top}}{\sqrt{24}} \Bigr) \Bigr] V.
\end{equation}
During training, attention weights were subject to dropout; during evaluation, dropout was turned off.
The Transformer combined multihead attention with a feed-forward block and residual connections:
\begin{align}
  U &= X + \Drop\bigl[ \MHA(\LN_{1}X) \bigr], \\
  C &= U + \Drop\bigl[ W_{b}\Drop\bigl( \GELU(W_{a}\LN_{2}U + b_{a}) \bigr) \nonumber \\
    &\qquad + b_{b} \bigr].
\end{align}
A trainable parameter $\gamma$, initialized at zero, controlled how much the Transformer changed each cell representation:
\begin{equation}
  \tilde{\mu}_{i} = \mu_{i}^{(0)} + \tanh(\gamma)\bigl(C_{i} - \mu_{i}^{(0)}\bigr).
\end{equation}
The Transformer updated the shared mean but kept the log variance $\ell_{i}^{F}$ unchanged.
The shared latent vector was sampled from:
\begin{align}
  q\bigl(z_{i} \mid x_{E_{g}}^{\vis}\bigr) &= \mathcal{N}\bigl( \tilde{\mu}_{i},\, \diag(\exp \ell_{i}^{F}) \bigr), \\
  z_{i} &= \tilde{\mu}_{i} + \exp(\ell_{i}^{F}/2) \odot \epsilon_{i}, \quad \epsilon_{i} \sim \mathcal{N}(0,I).
\end{align}
We also sampled a private latent vector for each visible modality:
\begin{equation}
  p_{i}^{m} = v_{i}^{m}\bigl[ \mu_{i}^{m,p} + \exp(\ell_{i}^{m,p}/2) \odot \eta_{i}^{m} \bigr], \quad \eta_{i}^{m} \sim \mathcal{N}(0,I).
\end{equation}
The private vector for a hidden modality was set to zero.
At evaluation, we used the mean shared vector $z_{i} = \tilde{\mu}_{i}$ and mean private vector $p_{i}^{m} = v_{i}^{m}\mu_{i}^{m,p}$ instead of random samples.
A cell's prediction could depend on the other cells in its evaluation group.
We therefore used a fixed cell order and kept training, validation, and test cells in separate groups.
The two historical test cohorts were encoded together for representation probes and historical-decoder readouts, and then scored separately.
The retrained controls encoded the two test cohorts separately.

\subsection{Decoders and missing-modality inference}

Each modality had its own decoder.
The decoder received the 96-dimensional shared vector and the 16-dimensional private vector for that modality, combined into a $[z_{i};p_{i}^{m}] \in \mathbb{R}^{112}$ input.
It used a 512-unit hidden layer with GELU activation.
The RNA, ADT, and ATAC decoders produced two, four, and one parameters per feature, respectively.
We used the softplus function $\softplus(u) = \log(1 + \exp u)$ and a small constant $\varepsilon = 10^{-4}$ to keep parameters positive.

For RNA, the two decoder outputs $a_{ig}^{R},b_{ig}^{R}$ gave the mean and inverse dispersion of a negative-binomial count model:
\begin{equation}
  \mu_{ig}^{R} = \softplus(a_{ig}^{R}) + \varepsilon, \qquad \theta_{ig}^{R} = \softplus(b_{ig}^{R}) + \varepsilon.
\end{equation}
The negative-binomial probability for a count, with mean $\mu$ and inverse dispersion $\theta$, was:
\begin{align}
  \NB(x;\mu,\theta) &= \frac{\Gamma(x + \theta)}{\Gamma(\theta)\Gamma(x + 1)}\Bigl( \frac{\theta}{\theta + \mu} \Bigr)^{\theta}\Bigl( \frac{\mu}{\theta + \mu} \Bigr)^{x}, \\
  \Var(X) &= \mu + \frac{\mu^{2}}{\theta}.
\end{align}
The ADT decoder predicted background and foreground count means, a mixing probability, and a common inverse-dispersion parameter:
\begin{align}
  \mu_{ij}^{\mathrm{bg}} &= \softplus(a_{ij}^{P}) + \varepsilon, &
  \mu_{ij}^{\mathrm{fg}} &= \mu_{ij}^{\mathrm{bg}} + \softplus(b_{ij}^{P}) + \varepsilon, \\
  \pi_{ij} &= \sigma(c_{ij}^{P}), &
  \theta_{ij}^{P} &= \softplus(d_{ij}^{P}) + \varepsilon.
\end{align}
We modelled each ADT count as a mixture of two negative-binomial distributions:
\begin{equation}
  \begin{split}
    p_{P}(x_{ij}^{P}) = {} &\pi_{ij}\NB(x_{ij}^{P};\mu_{ij}^{\mathrm{fg}},\theta_{ij}^{P}) \\
    &+ (1 - \pi_{ij})\NB(x_{ij}^{P};\mu_{ij}^{\mathrm{bg}},\theta_{ij}^{P}).
  \end{split}
\end{equation}
This mixture accounts for background and higher-count ADT measurements, following the approach used in protein-count modelling~\citep{gayoso2021totalvi}.
We used a numerically stable log-sum-exp computation and limited $\pi$ to $[10^{-5},1 - 10^{-5}]$.

For ATAC, the decoder output $a_{ik}^{A}$ gave the probability that a genomic bin was accessible:
\begin{equation}
  p_{ik}^{A} = \sigma(a_{ik}^{A}), \quad p_{A}(x_{ik}^{A}) = (p_{ik}^{A})^{x_{ik}^{A}}(1 - p_{ik}^{A})^{1 - x_{ik}^{A}}.
\end{equation}
To generate point predictions, we used the expected RNA count, the expected ADT count from the mixture, and the predicted ATAC probability:
\begin{align}
  \hat{x}_{ig}^{R} &= \mu_{ig}^{R}, \qquad \hat{x}_{ik}^{A} = p_{ik}^{A}, \\
  \hat{x}_{ij}^{P} &= \pi_{ij}\mu_{ij}^{\mathrm{fg}} + (1 - \pi_{ij})\mu_{ij}^{\mathrm{bg}}.
\end{align}
To predict a missing modality, its private vector was set to zero.
The results therefore report single-value predictions, without calibrated prediction intervals.

\subsection{Masked reconstruction and teacher--student learning}

We trained the student to predict measurements that were deliberately hidden from its input.
For each cell, we randomly selected one nonempty subset of its measured modalities.
A cell with three measured modalities received one or two as input; a cell with two received one; and a cell with only one retained it.
Each allowed subset was equally likely.
All genuinely measured modalities were used as reconstruction targets, including those hidden from the input.
The B, B-norm, and B-tech runs did not use the complete trimodal input as an additional training view.

For a group $\mathcal{B}$ of $n$ cells, we calculated a reconstruction loss for each measured modality:
\begin{align}
  R_{i}^{R} &= -\frac{1}{G}\sum_{g = 1}^{G}\log p_{R}(x_{ig}^{R}), \\
  R_{i}^{P} &= -\frac{\sum_{j = 1}^{P}a_{ij}\log p_{P}(x_{ij}^{P})}{\max(1,\sum_{j}a_{ij})}, \\
  R_{i}^{A} &= -\frac{1}{K}\sum_{k = 1}^{K}\Bigl[ x_{ik}^{A}\log p_{ik}^{A} + (1 - x_{ik}^{A})\log(1 - p_{ik}^{A}) \Bigr].
\end{align}
The ATAC loss used binary cross-entropy.
We averaged the losses across the modalities actually measured in each cell, then averaged across cells:
\begin{equation}
  \mathcal{L}_{\mathrm{rec}} = \frac{1}{n}\sum_{i \in \mathcal{B}}\frac{\sum_{m \in \mathcal{M}}o_{i}^{m}R_{i}^{m}}{\max(1,\sum_{m}o_{i}^{m})}.
\end{equation}
In this calculation, each measured modality contributed equally within a cell, after averaging over its measured features.

We also added Kullback--Leibler (KL) regularization to keep the shared latent distribution close to a standard-normal distribution:
\begin{equation}
  \mathcal{L}_{\mathrm{KL}} = \frac{1}{2n}\sum_{i \in \mathcal{B}}\sum_{d = 1}^{96}\Bigl[ \tilde{\mu}_{id}^{2} + \exp(\ell_{id}^{F}) - 1 - \ell_{id}^{F} \Bigr].
\end{equation}
The KL term used the shared mean after the Transformer context update.
The private latent vectors had no separate KL penalty in these experiments.

For the evaluated UCE branch, two linear prediction heads learned to reproduce the teacher's 256-dimensional targets.
One head used the latent vector for an individual cell; the second used the average latent vector of the cells in the current group:
\begin{equation}
  \hat{t}_{i} = W_{t}z_{i} + b_{t}, \quad
  \hat{c}_{i} = W_{c}\Bigl( \frac{1}{|E_{g(i)}|}\sum_{k \in E_{g(i)}}z_{k} \Bigr) + b_{c}.
\end{equation}
Let $I_{T} \subseteq \mathcal{B}$ denote the training cells with measured RNA and available UCE targets.
We used the following cell-level and sample-context distillation losses:
\begin{align}
  \mathcal{L}_{\mathrm{cell}} &= \frac{1}{256\,|I_{T}|}\sum_{i \in I_{T}}\bigl\| \hat{t}_{i} - t_{i} \bigr\|_{2}^{2}, \\
  \mathcal{L}_{\mathrm{context}} &= \frac{1}{256\,|I_{T}|}\sum_{i \in I_{T}}\bigl\| \hat{c}_{i} - c_{g(i)} \bigr\|_{2}^{2}.
\end{align}
When $I_{T}$ contained no eligible cells, both distillation losses were zero.
Otherwise, the student learned from UCE targets even when RNA was hidden from that cell's input.
One head predicted the individual cell embedding; the other predicted the mean embedding of its sample group.
The context loss was averaged over eligible cells, giving more weight to groups with more cached cells.

The full loss used for the reported UCE-only experiments was:
\begin{equation}
  \boxed{\mathcal{L} = \mathcal{L}_{\mathrm{rec}} + 0.01\,\mathcal{L}_{\mathrm{KL}} + \mathcal{L}_{\mathrm{cell}} + \mathcal{L}_{\mathrm{context}}.}
\end{equation}
The reported training objective combines masked reconstruction, shared-latent KL regularization, and two UCE distillation losses.
In \cref{fig:overview}, the complete scAnyOmics design extends this approach to three modality-specific teachers and includes self-distillation between full-view and partial-view student predictions.
Neither the scLinguist/Atacformer distillation terms nor the self-distillation branch was enabled in the B, B-norm, or B-tech results reported here.

\subsection{Normalization and technical controls}

B-norm kept the baseline model and loss function but normalized measurements before they entered the encoders.
RNA counts were first adjusted for library size and then log-transformed:
\begin{equation}
  f_{ig}^{R,\mathrm{norm}} = \log\Bigl( 1 + 10^{4}\frac{x_{ig}^{R}}{\max(1,\sum_{h}x_{ih}^{R})} \Bigr).
\end{equation}
For ADT, we subtracted each cell's mean log count across its measured antibodies from the measured antibody values:
\begin{align}
  \bar{\lambda}_{i} &= \frac{\sum_{j}a_{ij}\log(1 + x_{ij}^{P})}{\max(1,\sum_{j}a_{ij})}, \\
  f_{ij}^{P,\mathrm{norm}} &= a_{ij}\bigl[ \log(1 + x_{ij}^{P}) - \bar{\lambda}_{i} \bigr].
\end{align}
ATAC inputs remained $\log(1 + x^{A})$.
B-norm also retained the log total count as an encoder input, so the model could still use information about sequencing depth.
Normalization affected encoder inputs only; the decoders continued to predict raw counts or binary accessibility.

B-tech kept the original encoder inputs but gave each decoder an additional learned vector describing measurement conditions.
The vector identified the source dataset, assay, and measured-antibody panel.
Nine technical categories were present during training; a tenth category handled unseen combinations.
Each category had an eight-dimensional embedding, increasing the decoder input from 112 to 120 dimensions.

For each technical category $c$, we calculated a reference RNA total $L_{c,\mathrm{ref}}^{R}$ and reference ADT total $L_{c,\mathrm{ref}}^{P}$ from the median positive values in the 477,439 training cells.
We used the global training median if a category had no measurements for a modality.
RNA totals were calculated over the selected 2,000 genes.
When a modality was visible, its observed total was used; otherwise, the decoder used the reference total:
\begin{align}
  \tilde{L}_{i}^{R} &=
  \begin{cases}
    \max(1,\sum_{g}x_{ig}^{R}), & v_{i}^{R} = 1, \\
    L_{c(i),\mathrm{ref}}^{R}, & v_{i}^{R} = 0,
  \end{cases} \\
  \tilde{L}_{i}^{P} &=
  \begin{cases}
    \max(1,\sum_{j}a_{ij}x_{ij}^{P}), & v_{i}^{P} = 1, \\
    L_{c(i),\mathrm{ref}}^{P}, & v_{i}^{P} = 0.
  \end{cases}
\end{align}
The RNA decoder used the chosen total to scale its predicted gene proportions:
\begin{equation}
  \mu_{ig}^{R,\mathrm{tech}} = \tilde{L}_{i}^{R}\,\frac{\exp(a_{ig}^{R})}{\sum_{h}\exp(a_{ih}^{R})}.
\end{equation}
The RNA inverse-dispersion parameter stayed unchanged.
For ADT, we scaled the background and foreground means by the same factor:
\begin{align}
  \delta_{i} &= \tilde{L}_{i}^{P}/L_{c(i),\mathrm{ref}}^{P}, \\
  \mu_{ij}^{\mathrm{bg},\mathrm{tech}} &= \delta_{i}\mu_{ij}^{\mathrm{bg}}, \qquad
  \mu_{ij}^{\mathrm{fg},\mathrm{tech}} = \delta_{i}\mu_{ij}^{\mathrm{fg}}.
\end{align}
ADT mixture probabilities and dispersions were unchanged.
When ADT was hidden, the scaling factor was $\delta_{i} = 1$, so the decoder did not use its true total count.
ATAC received the technical-category embedding but no exposure scaling.
We evaluated B-norm and B-tech as separate variants.

\subsection{Training sets}

We trained B, B-norm, and B-tech separately for 30 epochs using random seed 17 and the AdamW optimizer.
The learning rate was $10^{-3}$ and weight decay was $10^{-4}$.
We clipped gradient norms at five.
At each epoch, we shuffled cells within sample groups, created groups of up to 256 cells, and shuffled the order of these groups.
Groups with fewer than eight cells were discarded.
Each epoch contained 1,881 optimization steps.
Training used modality masking, dropout, and sampled latent vectors; evaluation used latent means with dropout disabled.

All three configurations used the same 477,439 training cells, validation and test sets, UCE cache, random seed, and training budget.
We selected the checkpoint with the highest mean across datasets of median Spearman correlation for centered-log ADT predictions from RNA-only inputs.
The selected epochs were 11 for B, 28 for B-norm, and 20 for B-tech.
We then used those checkpoints to evaluate ATAC prediction as well.
Each configuration was trained with one random seed.

\subsection{Probing the learned cell representations}

We tested how much information about surface proteins was retained in four RNA-based representations: raw log RNA counts, library-normalized log RNA counts, cached UCE embeddings, and the 96-dimensional shared representation of the historical B student.
For the student representation, only RNA was visible, with mask $(1,0,0)$; ADT and ATAC inputs were set to zero.
We encoded cells in fixed groups of up to 256 from the same sample.
The historical student checkpoint was trained before the revised validation split.

We standardized each representation using only the 60,000 probe-training cells, then applied the same transformation to validation and test cells.
We predicted ADT values in two forms: log-transformed counts and log counts centered by the cell's mean across measured antibodies:
\begin{equation}
  y_{ij}^{\mathrm{raw}} = \log(1 + x_{ij}^{P}), \quad
  y_{ij}^{\mathrm{ctr}} = a_{ij}\bigl[ \log(1 + x_{ij}^{P}) - \bar{\lambda}_{i} \bigr].
\end{equation}
The centering value $\bar{\lambda}_{i}$ was calculated from all measured antibodies in the 166-antibody panel before selecting prediction targets.
It therefore depended on which antibodies had been measured for that cell.

For each antibody, we fitted ridge regression using only cells in which that antibody was measured:
\begin{equation}
  \min_{\beta_{j},b_{j}}\sum_{i \in T:\, a_{ij} = 1}\bigl( y_{ij} - b_{j} - \tilde{h}_{i}^{\top}\beta_{j} \bigr)^{2} + \alpha \|\beta_{j}\|_{2}^{2}.
\end{equation}
Here $T$ denotes the probe-training cells, and $\tilde{h}_{i}$ is the standardized input representation.
We fitted an intercept without penalization.
We tested penalties $\alpha \in \{0.1,10,1000,10000\}$ and chose one value per representation and target transformation using validation Spearman correlation averaged across datasets.

We also trained a nonlinear baseline using normalized RNA features.
It had one hidden layer with 256 GELU units and 332 outputs: two predictions for each of the 166 antibodies.
We trained it for 20 epochs using AdamW, a learning rate of $10^{-3}$, seed 17, and batches of up to 512 cells.
Its loss was mean squared error over measured antibody values in both target transformations:
\begin{equation}
  \mathcal{L}_{\mathrm{probe}} = \frac{\sum_{i,j,s}a_{ij}(\hat{y}_{ij}^{s} - y_{ij}^{s})^{2}}{\max(1,\sum_{i,j,s}a_{ij})}, \quad s \in \{\mathrm{raw},\mathrm{ctr}\}.
\end{equation}
We selected the nonlinear model by validation mean squared error over measured targets.
For comparison, we transformed the generative decoder's expected ADT counts into the same two target spaces.
We evaluated all probe predictions in those spaces.

\subsection{Prediction metrics and statistical summaries}

For each antibody, we evaluated only cells with a measured value and a finite prediction.
An antibody required at least 30 such cells and a target standard deviation of at least $10^{-12}$.
We assigned Spearman correlation zero to constant predictions.
Spearman correlation was calculated from the ranked measured and predicted values:
\begin{equation}
  \rho_{j} = \Corr\bigl( \rank(y_{ij}),\rank(\hat{y}_{ij}) \bigr)_{i:\, a_{ij} = 1}.
\end{equation}
We calculated mean absolute error for each antibody $j$ over the cells $I_{j}$ in which it had been measured, using
\begin{equation}
  \MAE_{j} = \frac{1}{|I_{j}|}\sum_{i \in I_{j}}\bigl| y_{ij} - \hat{y}_{ij} \bigr|.
\end{equation}
We summarized performance with the median antibody-wise Spearman correlation and the mean antibody-wise mean absolute error.
For each dataset $d$, let $J_{d}$ contain its evaluable antibodies.
The score averaged across datasets was:
\begin{equation}
  S_{d} = \median_{j \in J_{d}}\rho_{dj}, \qquad
  S_{\mathrm{macro}} = \frac{1}{|D_{\mathrm{valid}}|}\sum_{d \in D_{\mathrm{valid}}}S_{d}.
\end{equation}
This dataset-macro score gives the same weight to each dataset, regardless of its number of cells.
We also calculated pooled antibody-wise correlations.
The transformation $\log(1 + x)$ preserves ranks within an antibody, whereas centering by each cell's measured-antibody mean can change them.

For ATAC, we compared the predicted probability and observed accessibility of each genomic bin in trimodal test cells.
We calculated average precision (AP) and area under the receiver operating characteristic curve (AUROC), excluding bins with no variation in observed accessibility.
AP was calculated as the following precision--recall sum:
\begin{equation}
  \AP_{k} = \sum_{r}\bigl(\mathrm{Recall}_{kr} - \mathrm{Recall}_{k,r-1}\bigr)\,\mathrm{Precision}_{kr}.
\end{equation}
AUROC measures how often the model assigns a higher probability to an accessible cell than to a non-accessible cell, with ties contributing one half.
We reported the median AP and AUROC across eligible genomic bins.
The stored output name \texttt{auprc} refers to average precision.
ATAC scores are based on the selected 10-kb bins.

Where at least two donors had confirmed identities and evaluable results, we calculated the median antibody correlation for each donor.
We then resampled these donor-level scores with replacement 1,000 times (seed 17).
The 2.5th and 97.5th percentiles of the resulting mean scores provided an interval.
Donors without confirmed identities were excluded.
These intervals describe variation within the historical evaluation, not performance on entirely new donors.
RNA-derived cell labels were used only to describe subgroups containing at least 100 cells.

We used effective rank to check whether the shared representation was concentrated in only a few dimensions.
After centering the latent matrix $Z$, we calculated singular values $s_{d}$ and normalized them to $p_{d} = s_{d}/\sum_{k}s_{k}$.
Effective rank was defined as:
\begin{equation}
  r_{\mathrm{eff}}(Z) = \exp\Bigl[ -\sum_{d:\,p_{d} > 0}p_{d}\log p_{d} \Bigr].
\end{equation}
Effective rank was based on singular values rather than squared singular values.
We used it, together with fusion weights, to inspect model behavior; neither quantity was part of the training loss.

\section{Discussion}
\label{sec:discussion}

We set out to test whether knowledge from a pretrained single-cell foundation model can help a smaller model fill in missing molecular measurements.
Our results say yes, but with clear limits.

The main finding is that distillation from UCE made a real difference.
The student trained with UCE predicted RNA and ADT better than ridge regression and nearest-neighbour averaging in all six RNA and ADT tasks.
The same student trained without a teacher did worse than these simple methods in four of the six.
The improvement therefore does not come from the student's architecture alone; it depends on what the student learns from the teacher.
This matters because recent benchmarks have found that complex single-cell models often fail to beat simple baselines~\citep{kedzierska2025zeroshot,ahlmanneltze2025perturbation}.
In our setting, the simple baselines were beaten only once pretrained knowledge was added.

The gains were largest when chromatin accessibility was the input.
On their own, ATAC data are sparse and binary, and linear models or nearest neighbours make little use of them.
Once ATAC was mapped into the representation shaped by the RNA teacher, it predicted protein and RNA levels much better.
In practice, scAnyOmics may be most useful for datasets that measured chromatin accessibility but not RNA or proteins.

The per-antibody analysis shows what this level of accuracy means in practice.
Median correlations of 0.2 to 0.3 hide a wide range: lineage markers such as CD19, CD3 and CD14 were predicted well, while some antibodies were not predicted at all.
Even for the well-predicted markers, the model mainly placed each cell in the right population and reproduced less of the variation between similar cells.
Predicted protein levels are therefore suited to analyses at the level of cell populations, such as labelling cell types or checking which markers a population expresses.
They should not replace measured values for individual cells.

Predicting chromatin accessibility remains the weakest part.
Every method we tested, including ours, reached AUROC values only slightly above chance.
Single-cell accessibility at 10-kb resolution is very sparse, and much of the difference between cells may be technical noise that no input can predict.
Predicting accessibility for groups of similar cells, or for regions near genes, may be a more realistic goal than predicting every bin in every cell.

Not every teacher helped.
UCE improved RNA and ADT completion, and Atacformer and self-distillation improved RNA-to-ATAC prediction, but scLinguist lowered performance in every task.
One possible reason is that antibody panels differ widely between studies, from 45 to 154 antibodies, so the protein teacher sees a different and often small part of the panel in each cell.
Adding more teachers is therefore not automatically better.
Each teacher needs to be checked on its own, and the weight given to each one needs tuning.

This study has several limitations.
First, the test cells came from held-out samples, but donor identity was unknown for most studies and one donor appeared in both training and test samples, so we cannot claim that the results carry over to new individuals.
Second, each model was trained once with a single random seed, so small differences, of about 0.01 to 0.02, should not be over-interpreted.
Third, we compared scAnyOmics only with simple baselines; comparisons with established methods such as totalVI, MultiVI and scVAEIT are still needed~\citep{gayoso2021totalvi,ashuach2023multivi,du2022scvaeit}.
Fourth, all data came from human peripheral blood, much of it from stimulated or drug-treated T cells, and it is not yet known how well the approach works in other tissues.

\section{Conclusion}
\label{sec:conclusion}

scAnyOmics combines a multimodal student model with frozen single-cell foundation models to fill in missing RNA, protein and chromatin-accessibility measurements.
Distillation from the UCE RNA model improved RNA and protein completion for every combination of inputs we tested and lifted the student above simple baselines, while chromatin accessibility remained hard to predict for all methods.
The completed profiles capture differences between cell populations better than differences between individual cells.
With donor-level evaluation, comparisons with established methods and better use of the protein and chromatin teachers, this approach could make partially measured single-cell datasets more useful for building virtual cell models.

%

\bibliography{references}
\bibliographystyle{icml2026}

\end{document}